\documentclass[runningheads,a4paper]{llncs}

\usepackage{amssymb}
\usepackage{amsmath}
\usepackage{bold-extra}

\usepackage{listings}
\usepackage{graphicx}

\usepackage{url}
\urldef{\mailek}\path|ewan@inf.ed.ac.uk|
\urldef{\mailpp}\path|P.Pareti@sms.ed.ac.uk|

\newcommand{\ling}[1]{\textit{#1}}
\newcommand{\term}[1]{\textbf{#1}}
\newcommand{\FO}{\textsc{FO}}
\newcommand{\vago}{\textsf{VAGO}}

\begin{document}

\mainmatter

\title{Learning Vague Concepts for the Semantic Web}

\titlerunning{Learning Vague Concepts}

\author{Paolo Pareti
\and Ewan Klein}

\institute{School of Informatics, University of Edinburgh\\
10 Crichton Street, Edinburgh EH8 9AB, UK\\
\mailpp, \mailek\\
}

\maketitle

\begin{abstract}
  Ontologies can be a powerful tool for structuring knowledge, and
  they are currently the subject of extensive research. Updating the
  contents of an ontology or improving its interoperability with other
  ontologies is an important but difficult process. In this paper, we
  focus on the presence of vague concepts, which are pervasive in
  natural language, within the framework of formal ontologies. 
  We will adopt a framework  in which vagueness is captured via numerical 
  restrictions that can be automatically adjusted. Since
  updating vague concepts, either through ontology alignment or
  ontology evolution, can lead to inconsistent sets of axioms, we
  define and implement a method to detecting and repairing such
  inconsistencies in a local fashion.
\end{abstract}

\section{Introduction}

Historically, there has been a close relationship between ontologies
on the one hand, and glossaries, taxonomies and thesauri on the other
hand: although formal ontologies are expressed in a well-defined
formal language, many of the intuitions and terms used in ontologies
are derived from their natural language counterparts. Nevertheless, there is
an obvious mismatch between formal ontologies and natural language
expressions: vagueness is pervasive in natural language, but
is typically avoided or ignored in ontologies. 

Following standard usage, we will say that a concept is \term{vague}
when it admits borderline cases --- that is, cases where we are unable
to say whether the concept holds or fails to hold.\footnote{For recent
  overviews of the very extensive literature on vagueness, see
  \cite{VanDeemter:2010:NE,VanRooij:2011:VL}} The standard example
involves the adjective \ling{tall}. There are some people that we
regard as definitely tall, and others we regard as definitely short;
but people of average height are neither tall nor short. Notice that
the source of indeterminacy here is not lack of world knowledge: we
can know that John is, say, 1.80 metres in height, and still be
undecided whether he counts as tall or not. 

Rather than trying to capture vague expressions directly (for example,
by means of fuzzy logic), we will view vagueness as a property that
characterizes the definition of certain concepts over a
\emph{sequence} of ontologies deployed by an agent. While `ordinary'
concepts are treated as having a fixed meaning, shared by all users of
the ontology, we propose instead that the meaning of a vague concept is
\emph{unstable}, in the sense that the threshold on the scale of
height which distinguishes between being tall and not tall is
inherently defeasible.  

Is there any reason why we should care about ontologies being able to
express vagueness? As an example, consider FOAF \cite{foaf}, which is one of the most widely used
ontologies on the web.
One of FOAF's core predicates is \texttt{based\_near}. It is
instructive to read the commentary on this property:
\begin{quotation}\noindent
  The \texttt{based\_near} relationship relates two ``spatial things''
  (anything that can be somewhere), the latter typically described
  using the geo:lat / geo:long geo-positioning vocabulary\ldots

  We do not say much about what `near' means in this context; it is a
  `rough and ready' concept. For a more precise treatment, see
   GeoOnion vocab design discussions, which are aiming to produce a
  more sophisticated vocabulary for such purposes.
\end{quotation}
The concept is `rough and ready' in a number of senses: it is
undeniably useful; it is vague in that there are obviously borderline
cases; and it is also highly context-dependent. This latter issue is
addressed to a certain extent by the GeoOnion document
\cite{geo-onion} which is referenced above, but there is no
systematic attempt to get to grips with vagueness.

We have chosen to implement our approach in OWL 2
\cite{OWL2:2009:OLD}, since we are interested in targetting semantic
applications on the web, and OWL 2 is sufficiently expressive for our
purposes while offering efficient reasoners such as Pellet
\cite{Sirin:2007:PPO} and HermiT \cite{Motik:2007:ORD}. Since the
relevant aspects of OWL 2 can also be expressed more compactly in
Description Logic (DL) \cite{Baader:2007:DL}, we will use the latter
as our main vehicle for representing ontologies.

In this paper, we will start off (\S1) by considering how to
accommodate vague concepts into a framework such as Description
Logic, and we will also situate the discussion within the wider perspective of
ontology evolution and ontology alignment. Then \S2 presents the
architecture of the \vago\ system, which treats vague concepts as
defeasible; that is, able to be updated when new information is
acquired. \S3 describes and discusses a number of experiments in which the
implemented \vago\ system runs with both artificial and real-world
data. Finally, \S4 provides a conclusion.

\section{Representing and Updating Vague Concepts}

\subsection{Gradable Adjectives and Measures}
\label{sec:gradable-adjectives}

Adjectives such as \ling{tall}, \ling{expensive} and \ling{near} are
often called \term{gradable}, in that they can be combined with degree
modifiers such as \ling{very} and have comparative forms
(\ling{taller}, \ling{more expensive}). As a starting point for
integrating such predicates into Description Logic,
consider the following degree-based semantic representation of the predicate
\ling{expensive}
\begin{equation}
  \label{eq:4}
  \mathit{expensive} \equiv \lambda x.\exists
  d[\mathbf{C}(d) \wedge \mathbf{expensive}(x) \preceq d]
\end{equation}
Here, ``\textbf{expensive} represents a measure function that takes an
entity and returns its cost, a degree on the scale associated with the
adjective'' \cite{Kennedy:2005:SSS}, p.349.
The predicate \textbf{C} is a contextually-given restriction
which determines the threshold for things that are definitely
expensive. Thus, an object will be expensive if its cost is greater
than the threshold $d$.
The relation $\mathbf{expensive}$ resembles a datatype property in
OWL, associating an individual with a data value. In DL, we could
introduce a concept $\mathsf{Expensive}$ and constrain its
interpretation with a datatype restriction of the following kind
\cite{Magka:2010:TED}, where $X$ is a variable whose role we will
discuss shortly:
\begin{equation}
  \label{eq:5}
  \mathsf{Expensive} \sqsubseteq \exists \mathsf{hasMeasure}.(\geq,X)
\end{equation}
In \cite{Baader:1991:SIC}, an expression such as $(\geq,200)$ is
called a predicate name, and corresponds to an abstraction over a
first-order formula, e.g., $\lambda x.x \leq 200$.

To gain a more general approach, we will adopt the approach to
adjectives proposed in \cite{Amoia:2006:ABI} and make the predicate's scale (in
this case, the cost) explicit:
\begin{equation}
  \label{eq:6}
  \mathsf{Expensive} \equiv \exists \mathsf{hasProperty}.(\mathsf{Cost} \sqcap \exists \mathsf{hasMeasure}.(\geq,X))
\end{equation}
However, we are left with a problem, since we still need some method
of to provide a concrete value in place of $X$ in particular contexts
of use.  We will regard $X$ as a metavariable, and to signal its
special function, we will call it an \term{adaptor}.  Any DL axiom
that contains one or more adaptors is an \term{axiom
  template}. Suppose $\phi[X_1, \ldots X_n]$ is an axiom template,
$\mathcal{X} = \{X_1, \ldots X_n\}$ is the set
of adaptors in $\phi$ and $T^{\mathcal{D}}$ is a set of datatype
identifiers. A \term{assignment} $\theta: \mathcal{X} \mapsto T^{\mathcal{D}}$ is
a function which binds a datatype identifier to an adaptor.
We write  $\phi[X_1, \ldots X_n]\theta$ for the result of applying the assignment
$\theta$ to $\phi[X_1, \ldots X_n]$. For example, if  $\phi[X]$ is the
axiom template in (\ref{eq:5}), then $\phi[X]\{X \leftarrow 200\}$ is 
$\mathsf{Expensive} \sqsubseteq \exists \mathsf{hasMeasure}.(\geq,200)$.

\subsection{Delineations}
\label{sec:vague-concepts}

Gradable adjectives typically come in pairs of opposite polarity; for
example, the negative polarity opposite of \ling{expensive} is
\ling{cheap}, which we can define as follows:
\begin{equation}
  \label{eq:7}
  \mathsf{Cheap} \equiv \exists \mathsf{hasProperty}.(\mathsf{Cost} \sqcap \exists \mathsf{hasMeasure}.(\leq,X'))
\end{equation}
Should the value of $X'$ in (\ref{eq:7}) --- the upper bound of
definitely cheap --- be the same as the value of $X$ in  (\ref{eq:6})
--- the lower bound of definitely expensive? On a partial semantics
for vague adjectives, the two will be different. That is, there be
values between the two where things are not clearly expensive or cheap.
One problem with partial valuation is that if $C(x)$
is neither true nor false for some $x$ then plausible treatments of logical
connectives would give the same undefined value to $C(x) \wedge \neg
C(x)$ and $C(x) \vee \neg C(x)$. However, many logicians would prefer
these propositions to retain their classical values of true and false
respectively. To address this problem, Fine \cite{Fine:1975:VTL} and
Kamp \cite{Kamp:1975:TTA} proposed the use of \term{supervaluations},
namely valuations which make a partial valuation more precise by
extending them to a total valuation of the standard kind. In place of
formal details, let's just consider a diagram:
\begin{figure} [h]
\label{supervals}
       \centering
        \includegraphics[width=0.75\textwidth]{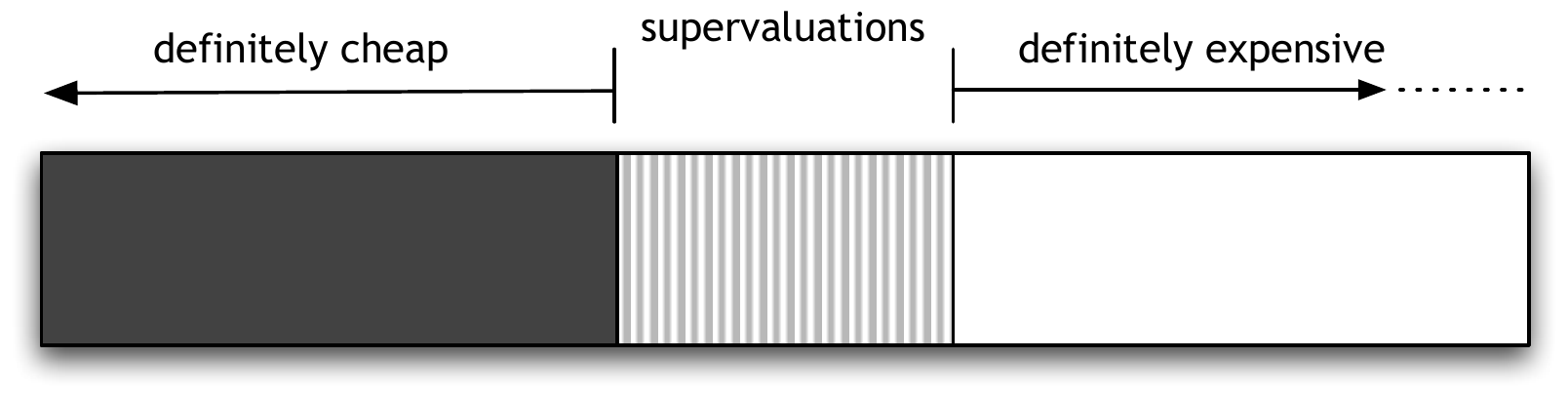}
\caption[]{Delineations of a Vague Concept}
\end{figure}
That is, each supervaluation can be thought of as way of finding some
value $v$ in the `grey area' which
is both an upper bound threshold for \ling{cheap} and a lower bound
threshold for \ling{expensive}.

We adopt a model of vagueness in which vague concepts
\emph{do} receive total interpretations, but these are to be regarded as
similar to supervaluations, in the sense that there may be multiple
admissible delineations for the concept. The particular delineation
that is adopted by an agent in a specific context can be regarded as
the outcome of learning, or of negotiation with other
agents. Consequently, on this approach, vague concepts differ from crisp concepts by
only virtue of their instability: a vague concept is one where the
threshold is always open to negotiation or revision. As we have
already indicated,
the defeasibility of threshold is not completely open, but is
rather restricted to some borderline area; however, we will not
attempt to formally capture this
restriction here.\footnote{For more discussion, see
  \cite{Lehmann:1994:EYR,Burato:2009:LMN}.}

\subsection{Ontology Evolution and Ontology Alignment}
\label{sec:ontology-evolution}

As part of a learning process, an agent should be prepared to update
the value of adaptors occurring in concepts in its ontology. New
values can be learned as a result of interaction with other
agents---corresponding to \term{ontology alignment}
\cite{Euzenat:2007:OM,Choi:2006:SOM}, or as a result of updating its
beliefs about the world---corresponding to \term{ontology evolution}
\cite{Flouris:2008:OCC}.  We will examine two important issues. The
first concerns the question of how to automatically update vague
concepts in an ontology as a result of learning new information
\cite{Zablith:2009:OEE}. The second, closely related issue, is how to
ensure that an ontology is still consistent after some of its axioms
have been modified \cite{Khattak:2010:AEO}.

Let's assume we have two ontologies $O_1 = (S, A_1)$ and $O_2 = (S,
A_2)$ which share a signature $S$ but have differing axiom sets $A_1$
and $A_2$. Axioms will consist of concept inclusions $C \sqsubseteq
D$, concept assertions $C(a)$ and role assertions $r(a, b)$. Suppose
$A_1$ contains the following axioms:
\begin{eqnarray}
\mathsf{LegalAdult(Jo)}\\
\label{eq:100}\mathsf{LegalAdult} \equiv \mathsf{Person} \sqcap \exists\mathsf{hasAge}.(\geq, 18) 
\end{eqnarray}
We now want to update $O_1$ with the axiom set $A_2$, which happens to contain the
following:
\begin{equation}
  \mathsf{hasAge(Jo, 17)}
\end{equation}
How do we deal with the ensuing inconsistency? The core of our
proposal involves identifying the numerical restriction in
(\ref{eq:4}) as the value of an adaptor parameter, and therefore open
to revision. That is, like other approaches to Ontology Repair, we
need to identify and modify axioms that are responsible for causing an
inconsistency. However, we localize the problem to one particular
component of axioms that provide definitions for vague concepts. In
this example, the inconsistency could be solved by changing the value
$18$ used in the definition of a \textsf{LegalAdult} to the value of
$17$.

\section{System Architecture}
\label{sec:system-architecture}

\subsection{Overview}

As described earlier, vagueness is captured by the `semantic instability'
of certain concepts. This can be seen as an extreme kind of
defeasibility: the threshold of a vague concept has a propensity to
shift as new information is received.  We have developed a
a computational framework called \vago\ in which
learning leads to a change in the extension of vague concepts via the
updating of adaptors. This is the only kind of ontology update that we
will be considering here.

The first input to \vago\ is the ontology $O = (S, A)$ that will potentially be
updated.  We call this the \term{original ontology} to distinguish it
from the \term{updated ontology} which is the eventual output of the
system. The second input is a set $A_{t}$ of axioms, here called
\term{training axioms}, that will be used to trigger an update of the
original ontology. $A_{t}$ is assumed to be part or all of the axiom
set of some other ontology $O' = (S, A')$, and uses the same signature
$S$ as $O$.

A schematic representation of \vago 's architecture is shown
in Fig\,2.
\begin{figure} [htb]
\label{schemaArchitecture}
       \centering
        \includegraphics[width=0.75\textwidth]{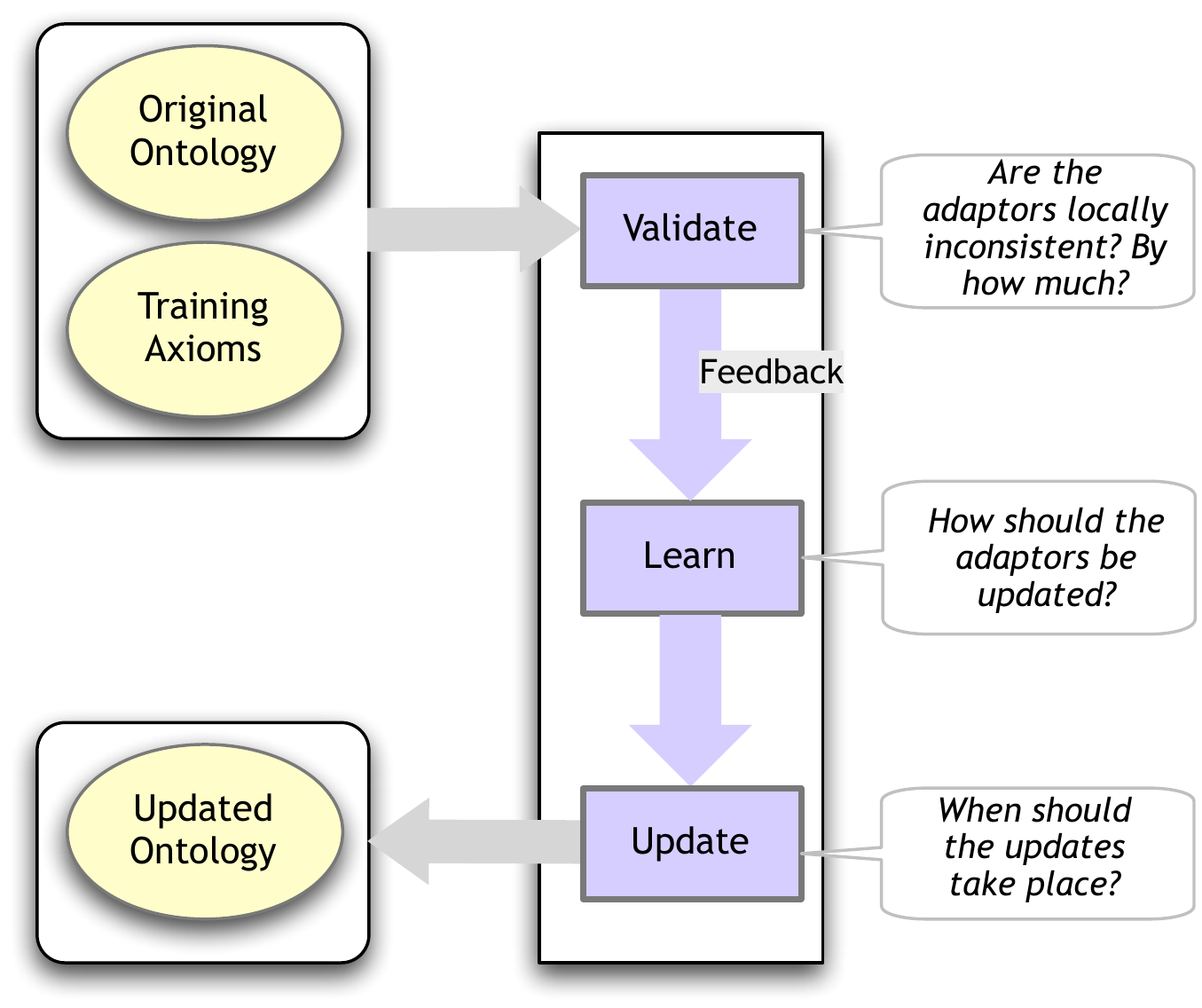}
\caption[]{Diagram of System Architecture}
\end{figure}

Given the original ontology and
the training axioms as inputs, the framework will output an updated version
of the ontology. The whole process of computing this output can be
divided into three main phases: validation, learning
and update.

The goal of the validation phase is to extract diagnostic feedback from the 
training axioms. This feedback should provide information about the adaptors
used in the original ontology. In particular, it should state whether an adaptor
is responsible for an inconsistency and if so, propose an alternative
value for the adaptor to remove the inconsistency.

The purpose of second phase, namely learning, is to determine how the
values of adaptors should be updated, given the diagnostic feedback
extracted during validation. These updates will in turn form the input
of the third phase, which controls in detail how the original ontology
should be modified to yield the updated ontology.

\subsection{Validation Phase}

Validation is the first phase of \vago\ and examines the training
axioms to produce a number of diagnostic \term{feedback objects}
(\FO s for short). These will contain a compact representation of all the
information required for the subsequent learning phase. Let's suppose
that in the original ontology, it is asserted that only people over
the age of 18 are legal adults, where 18 is the value of adaptor $X$.  
In DL this assertion could be
represented by the following axiom template: 
\begin{equation}
\label{eq:adult}
\mathsf{Adult} \equiv \mathsf{Person}\,\sqcap\,\exists \mathsf{hasAge}.(\geq,\, X)  
\end{equation}
with the adaptor $X$ instantiated to the value of 18. 
\begin{table}
\caption{Example of feedback objects for adaptor $X$}
\label{feedbackExampleTable}
\centering
\begin{tabular}{lll}
\hline
\multicolumn{1}{c}{} & \FO-1 & \FO-2\tabularnewline
\hline 
Adaptor identifier & $X$ & $X$\tabularnewline
\hline 
Was the adaptor correct? & False & True\tabularnewline
\hline 
Current value of the adaptor & 18 & 18\tabularnewline
\hline 
Required value for the adaptor & 16 & 26\tabularnewline
\hline 
\end{tabular}
\end{table}
Table\,1

shows the feedback that the
validation phase would output after examining the following
training axioms:
\begin{eqnarray}
\label{ax:a1}\mathsf{LegalAdult(John)}\\
\label{ax:a2}\mathsf{hasAge}(\mathsf{John}, 16)\\
\label{ax:a3}\mathsf{LegalAdult(Jane)}\\
\label{ax:a4}\mathsf{hasAge}(\mathsf{Jane}, 26)
\end{eqnarray}
In this example, training axioms (\ref{ax:a1}) and (\ref{ax:a2})
are incompatible with the definition of \textsf{LegalAdult} in the original
ontology, generating an inconsistency. More specifically, under the
assignment $X \leftarrow 18$, the set consisting of (\ref{eq:adult})
and axioms (\ref{ax:a1}), (\ref{ax:a2}) is inconsistent, and in such a
case we shall say that  $X \leftarrow 18$ (or more briefly, just $X$)
is \term{locally inconsistent}. However, this
inconsistency can be removed by assigning a new value to $X$; more specifically,
no inconsistency would arise for $X$ holding a value less than or equal
to $16$. The optimal new value is one that 
removes the inconsistency with the least modification to the current
value of $X$, which in this case is 16.

How is it possible to automatically determine, among all the possible
values of $X$, the value $v$ that will remove the
inconsistency while differing least from the current value of $X$?
Fortunately, if $v$ exists, it is straightforwardly recoverable from
training axioms.  It is therefore sufficient
to consider only a small set of possible candidate values for the
adaptor. Given a set of inconsistent axioms (possibly minimal) and an
adaptor $X$, algorithm \ref{alg_computePossibleValues} shows how to
extract this candidate set.

\begin{figure}[t]
  \centering
\begin{lstlisting}[caption={Compute all candidate values for set $A$ of inconsistent axioms and adaptor $X$},label=alg_computePossibleValues,mathescape=True]
computeAlternativeValues(parameters: inconsistent_axioms, adaptor)
  values $\leftarrow$ empty set
  data_relations $\leftarrow$ set of relations in inconsistent_axioms restricted by the adaptor on value of their target
  cardinality_relations $\leftarrow$ set of relations in inconsistent_axioms restricted by the adaptor in their cardinality
  all_individuals $\leftarrow$ set of individuals in inconsistent_axioms
  foreach individual in all_individuals do
    foreach r in data_relations do
      data_values $\leftarrow$ set of all values that individual is related to by relation r
      values $\leftarrow$ values + all data_values
     end
     foreach r in cardinality_relations do
       cardinality $\leftarrow$ number of relations r that the individual has       
       values $\leftarrow$ values + cardinality
     end
  end
  return (values)
\end{lstlisting}
\end{figure}

Given the set  $V$ of values computed by Algorithm \ref{alg_computePossibleValues}
for a set  $A$ of inconsistent axioms and an adaptor $X$, if there
is an assignment $X \leftarrow v$ which will remove the inconsistency
from axioms $A$, then $v$ is either included in $V$, or it is the
immediate predecessor or successor of a value in $V$.\footnote{We 
define $b$ to be the \term{immediate successor} of $a$ if
$a<b$ and there is no other value $c$ such that $a<c<b$; the immediate
prececessor is defined in a parallel manner.}
For each value $v \in V$, it is necessary to consider also the
first successor 
and first predecessor to deal with both strict and non-strict
inequalities in numerical constraints.

\subsection{Learning Phase}

The learning phase is responsible for computing the updates that the
original ontology should adopt, given the feedback objects extracted
from the training axioms.  The feedback objects are intended to
provide the evidence necessary to justify a change in the adaptors.

If an adaptor assignment was found to be locally inconsistent, then it reasonable to assume
that there is evidence to support its change. More specifically, given
feedback to the effect that assignment $X \leftarrow v_0$ is locally
inconsistent whereas $X \leftarrow v_1$ is not, then there is evidence
to support a new assigment $X \leftarrow v_2$, where $v_2 = (v_1-v_0)$.

The validation phase can discover different pieces of evidence
supporting different values $[v_{1},v_{2},...,v_{n}]$ for the same
adaptor $X$. We will assume that the update to be computed should
be the arithmetic mean of all these possible values.  If the information contained
in the training axioms is subject to noise, it will be
desirable to reduce the importance of new values that are 
far from the mean $\bar{v}$. For this purpose, we use
a sigmoid function $l:\, \mathbb{R} \mapsto[0,1]$ to reduce exponentially
the importance of a candidate value $v$ the further it is from the
mean $\bar{v}$.
Values far from the mean
will be scaled by a factor close to 0
(e.g., $l(\infty)=0$) while those close to the
mean will be scaled by a factor close to 1 (e.g., $l(0)=1$).
Given the distance $x$ from the mean and the standard deviation $\delta$
over the set $V$ of candidate values, a plausible definition for the function $l$
might be the following 
(where $q$ and $b$ are free parameters):
\[
l(x)=\frac{1}{\left(1+(\delta/q)e^{-b(x-2\delta)}\right)^{q/\delta}}
\]
After these additional considerations, the update $u$ can be
computed analytically using the following formula: 
\[
u=\frac{1}{n}\left(\sum_{i=1}^{n}v_{i}\, l(|v_{i}-\bar{v}|)\right)
\]

\subsection{Implementation}
\label{sec:system-implementation}

We have built a Java implementation of \vago\ using the OWL API version
3.2.3 \cite{OWLAPI} to manipulate OWL 2 ontologies.\footnote{The
  code for the implementation is available at
  \url{http://bitbucket.org/ewan/vago/src}.}  Adaptors are identified
and modified by accessing the XML/RDF serialization of the ontology.
The operations on the XML data make use of the JDOM API version 1.1.1
\cite{JDOM}.  Additional information has to be added to the ontology
in order to represent adaptors. To preserve the functionality of the
OWL ontologies, any additional information required by \vago\ can
be encoded as axiom annotations, or as attributes in the RDF/XML
serialization of the ontology. As a result, this information will be
transparent to any reasoner and it will not change the standard
semantics of the ontology.

Adaptors will be identified in OWL ontologies using special labels.
More specifically, if a value $v$ used in axiom $A$ is labeled with
an unique identifier associated with adaptor $X$, then it is possible
to say that $X$ is currently holding the value $v$ and that the
axiom $A$ is \emph{dependent} on the adaptor $X$. When the value
of adaptor $X$ is updated to a new value $z$, then
the value $v$ in axiom $A$ will be replaced with the new value
$z$. If multiple axioms of an ontology are dependent on adaptor $X$,
then all their internal values associated with $X$ will change accordingly.

\section{Experiments}
\label{sec:evaluation}

\subsection{Experiments using Artificial Data}

The first evaluation of \vago\ uses an ontology describing persons
and simple relations between them. The most important definitions
in this ontology are the following:
\begin{itemize}
\item $\mathsf{Person}$: a general class representing a human being;
\item $\mathsf{Minor} \equiv \mathsf{Person}\;\sqcap\;\exists
 \mathsf{hasAge}.(<,X_{1})$: a class representing a young person
 (defined as a person under the age of $X_{1}$);
\item $\mathsf{LegalAdult} \equiv \mathsf{Person}\;\sqcap\;\exists
 \mathsf{hasAge}.(\geq,X_{1})$: a class representing an adult (defined
 as a person of age $X_{1}$ or older);
\item $\mathsf{BusyParent} \equiv \mathsf{Person}\;\sqcap\; \geq\!
  X_{2}\mathsf{parentOf}.\mathsf{Minor}$: a class representing the vague
 concept of a busy parent (defined as a person with at least $X_{2}$
  young children);
\item $\mathsf{RelaxedParent} \equiv \mathsf{Person}\;\sqcap\;\exists
  \mathsf{parentOf}.\mathsf{Person}\;\sqcap\;\neg \mathsf{BusyParent}$: a class representing the
  vague concept of a relaxed parent (defined as a parent that is not
  busy);
\item $\mathsf{hasAge}$: a functional data property with domain $\mathsf{Person}$ and range
integer values;
\item $\mathsf{parentOf}$: an object relation between two
  $\mathsf{Person}$s, a parent and a child.
\end{itemize}
The training axioms used in each iteration are produced
automatically by generating instances of the above mentioned classes,
and the relations between them. Since the data is produced
automatically, it is possible to know the exact value that the
adaptors should have, namely the value used while producing the
data.
\begin{figure}
\centering
\includegraphics[clip,angle=-90,width=1\columnwidth]{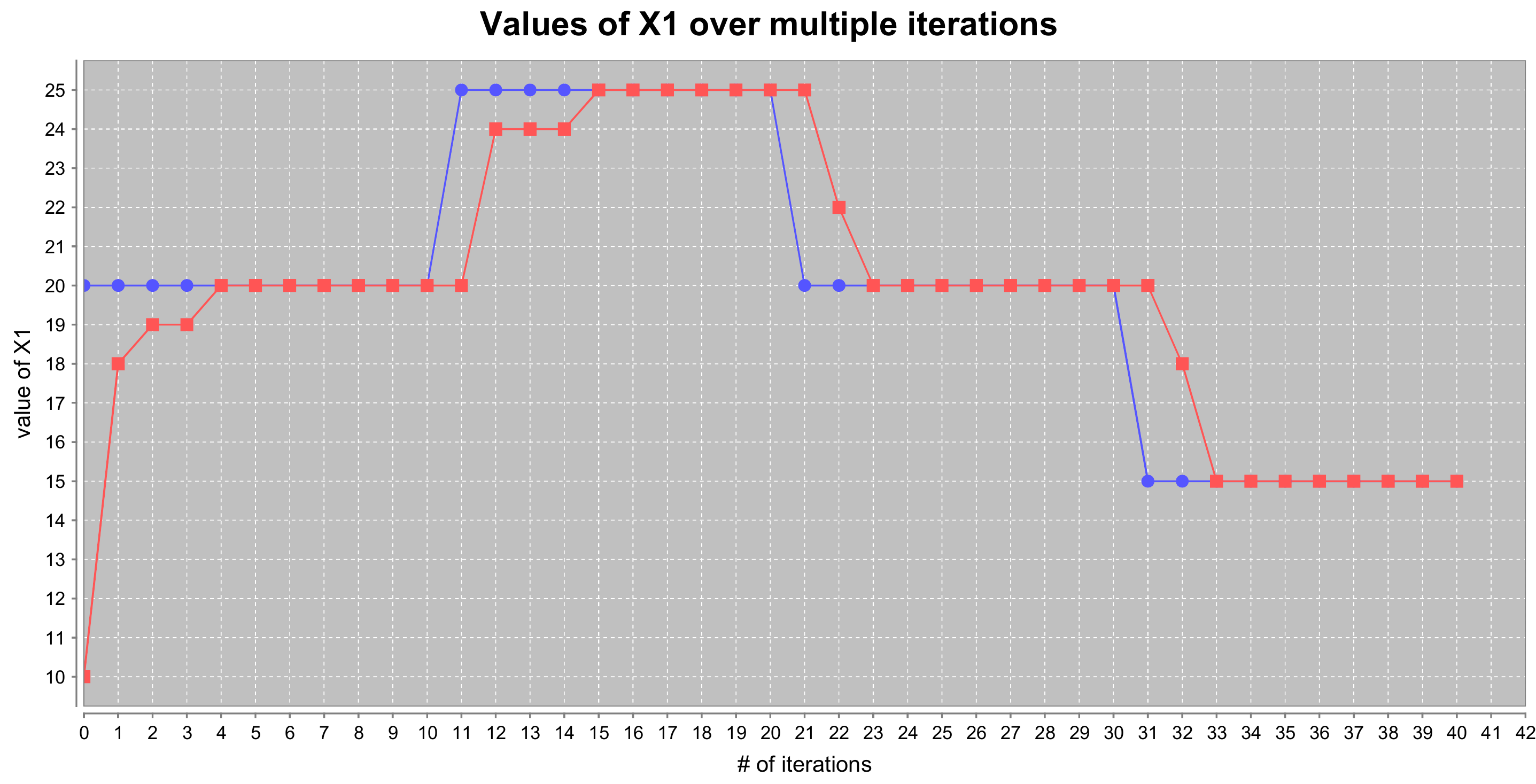}
\caption{Plot of the values of adaptor $X_{1}$ across 40 iterations of the
system. The red squares indicate the value computed by
the system, the blue circles indicate the correct value for that iteration.}
\label{PLOTX1-noNoise}
\end{figure}

In the first scenario, the system tries to adjust the adaptor $X_{1}$
that defines the threshold between the concepts $\mathsf{Minor}$ and
$\mathsf{LegalAdult}$ by examining a number of individuals described
in the training axioms. Forty sets of training axioms are generated,
each one containing information about thirty persons. Whether these
individuals are labelled as $\mathsf{Minor}$ or $\mathsf{LegalAdult}$
depends on their age (randomly determined by the generation algorithm)
and on the correct value that $X_{1}$ should have in that iteration.
The correct value for this adaptor changes every 10 iterations to test
whether the system can adapt to changes in the environment. The
results of this simulation are shown in Fig.\,\ref{PLOTX1-noNoise}.
It can be seen that under these conditions the adaptor $X_{1}$ quickly
converges to a consistent value.

\begin{figure} 
\centering
\includegraphics[bb=0bp 0bp 394bp 842bp,angle=-90,width=1\columnwidth]{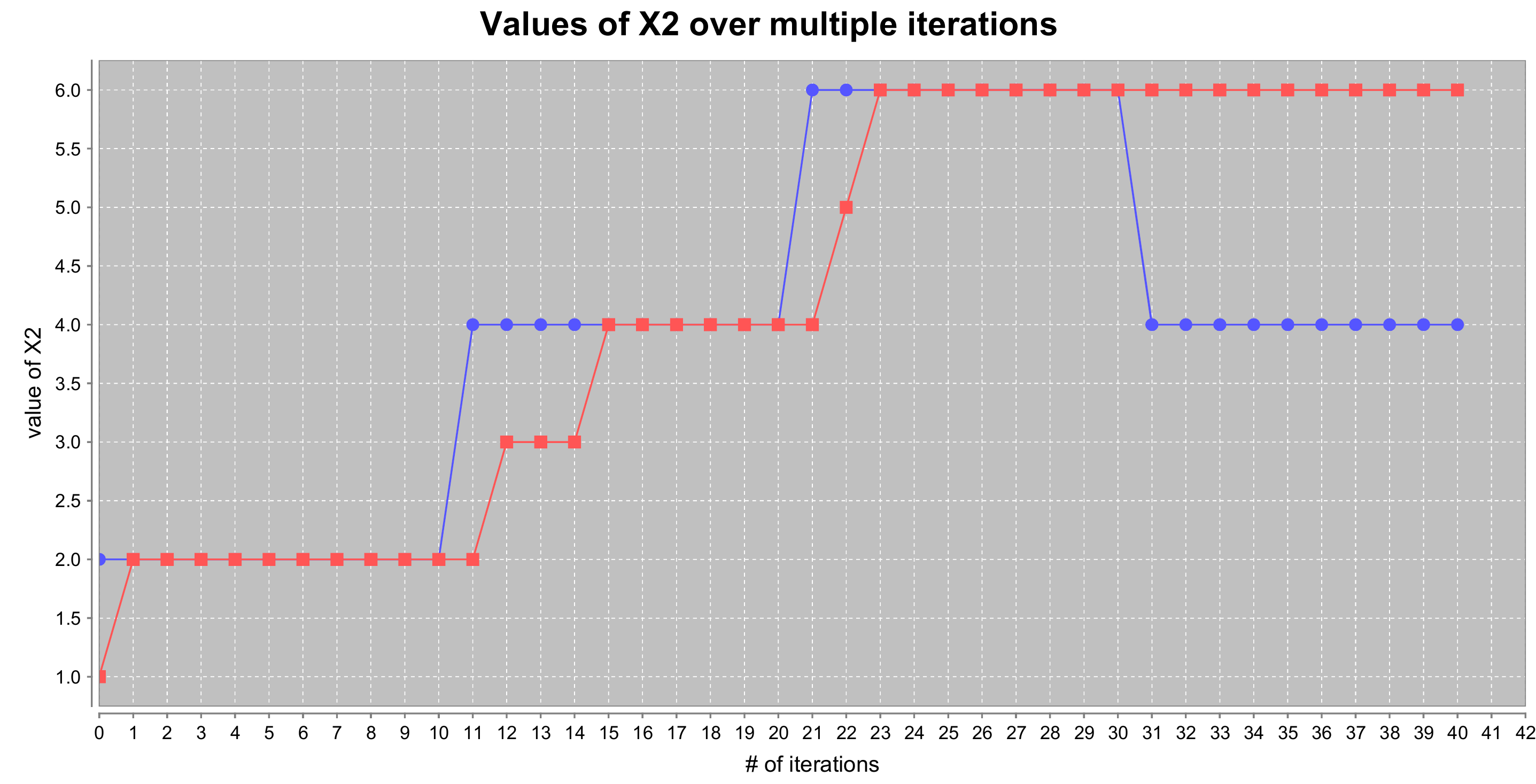}
\caption{Plot of the values of adaptor $X_{2}$ across 40 iterations of the
system.~The squares indicate the value computed by the system,
the circles indicate the target value for that iteration.}
\label{PLOTX2}
\end{figure}

The second scenario is concerned with the evolution of the
adaptor $X_{2}$, which restricts the cardinality of a relation.
In each of the 40 iterations of the system, a set of training axioms
is generated so that the value of $X_{2}$ in that iteration is varied randomly.
Each set of training axioms contains information about
ten instances of the $\mathsf{BusyParent}$ class, such that they are 
$\mathsf{parentOf}$ at least  $X_{2}$ children.
It also contains ten instances
of the class $\mathsf{RelaxedParent}$, such that they are
$\mathsf{parentOf}$ less than $X_{2}$ children.

Fig.\,\ref{PLOTX2} illustrates the result of the simulation in this
second scenario, and shows that an adaptor restricting a cardinality
(in this case $X_{2}$) can converge to a consistent value as long
as the convergence involves an increase in the value. If, instead, the convergence
\emph{reduces} the value of the adaptor (as in the
last ten iterations of this simulation), the adaptor will not change.
For this reason, $X_{2}$ is not able to adapt to the last change
in the simulation, namely reducing its value from 6 to 4. 
The inability to reduce the value of $X_{2}$, which might seem undesirable
from a practical point of view, is a logical consequence of the Open
World Assumption \cite{OpenClosedWorld} made by the reasoner used.

\subsection{Experiments with Web Data}

This second evaluation deals with a hypothetical scenario where the
user of \vago\ is a company that owns bookshops in cities around
the world. This company is interested in developing an automatic system
to discover cities where it might be profitable to open a new bookshop by
providing a sufficient definition for the concept `profitable place'.
We will here make the simplified assumption that a city centre will be classified 
as profitable place to open a business if there are few competitor bookshops nearby.
In the centre of a city, potential customers of books are assumed to reach a bookshop
on foot. Therefore in this context the concept `near' should
be interpreted as within walking distance. However, even after this clarification,
two concepts remain underspecified. How many bookshops should there
be in a city centre to be able to say that
they are ``too many''? And how many meters away should an object be to count as ``near''? 

These vague concepts are defined in an OWL ontology using two adaptors.
The first one, $C$, determines the maximum number of nearby
bookshops that a place can have while still being considered profitable. The second
one, $D$, determines the maximum number of meters between two places
that are considered near to each other.

The most important definitions contained in the original ontology
used in this simulation are the following:
\begin{itemize}
\item An instance of class $\mathsf{Distance}$ should be related to
  two $\mathsf{SpatialThing}$ (the places between which the distance
  is computed) and to one integer (the meters between the two places):\\
  $\mathsf{Distance}\equiv\,=2\,\mathsf{distBetween}.\mathsf{SpatialThing}\,\sqcap=1\,\mathsf{distMeasure}$
  \vspace{3 mm}
\item To be considered near, two places should have a $\mathsf{CloseDistance}$
  between them
  (a $\mathsf{Distance}$ which measures no more than $D$ meters):\\
  $\mathsf{CloseDistance}=\mathsf{Distance}\;\sqcap\;\exists
  \mathsf{distMeasure}.(\leq, D)$ \vspace{3 mm}
\item A $\mathsf{ProfitablePlace}$ is a place that has no more than $C$ bookshops nearby.
In DL, it could be expressed as follows:\\
$\mathsf{ProfitablePlace} \equiv \mathsf{SpatialThing}\;\sqcap\;\\
\;\leq\; C\; \mathsf{hasDistance}.(\mathsf{CloseDistance}\;\sqcap\;\exists \mathsf{distBetween}.\mathsf{bookshop})$
\end{itemize}

In order to learn the proper values to give to the adaptors $C$ and
$D$, a series of training axioms is fed to the system. 
More specifically, the city centres of the 30 largest city of the
United Kingdom are classified
as a $\mathsf{ProfitablePlace}$ and then additional information about each
city is extracted using web data. This additional information
describes places (such as bookshops) and the distances between them. 
To begin with, the location of a city centre is extracted from DBpedia \cite{DBpedia}.
A number of places around that city centre are then obtained from the
Google Places API \cite{GooglePlaces}. Some of these will be
already classified Google as bookshops
or as having a {\textsf{CloseDistance} between them.
The distances between them are then computed by the Google Distance
Matrix API \cite{GoogleDistance}. 
The resulting information was subject to noise;
for example, several distances measuring more than ten
kilometers were classified as $\mathsf{CloseDistance}$. The sigmoid function
used in the learning phase reduced the effect that values greatly differing from the mean have
on the evolution of the adaptors.

Fig.\,\ref{PLOT_d} shows the evolution of adaptor $D$ concurrent with
the evolution of the adaptor $C$ shown in Fig.\,\ref{PLOT_c}. The
ontology produced after the thirty iterations of this simulation
defines a place to be profitable if there are no more than four bookshops
within 1,360 meters distance.

A possible way to determine the correct value for the adaptor $C$ is
to consider the average number of bookshops near the city centres plus
its standard deviation across the iterations (considering just the
information contained in the training axioms).  This value is found to
be equal to 2.49.  In a similar way the average measure of a
$\mathsf{CloseDistance}$ plus the standard deviation is calculated as 1,325.
Assuming those values as the correct ones, the final value computed by
the system for the adaptor $D$ differs from the correct value by 4\%
of the standard deviation. The final value for adaptor $C$ differs
from the correct value by 162\% of the standard deviation.

\begin{figure} [htb]
\centering
\includegraphics[angle=-90,width=1\columnwidth]{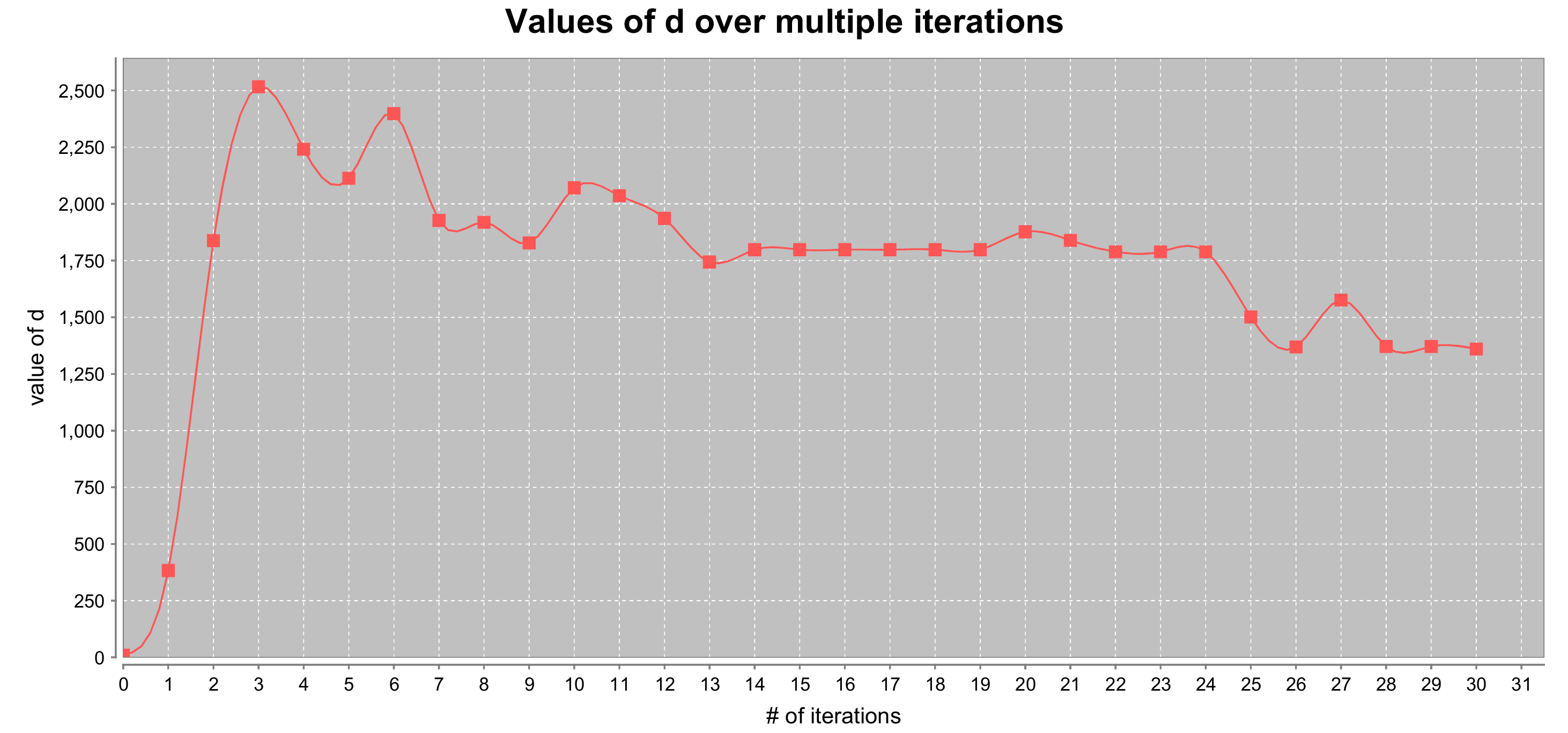}
\caption{Plot of the values of adaptor $d$ across 30 iterations of the system}
\label{PLOT_d}
\end{figure}

\begin{figure} [htb]
\centering
\includegraphics[angle=-90,width=1\columnwidth]{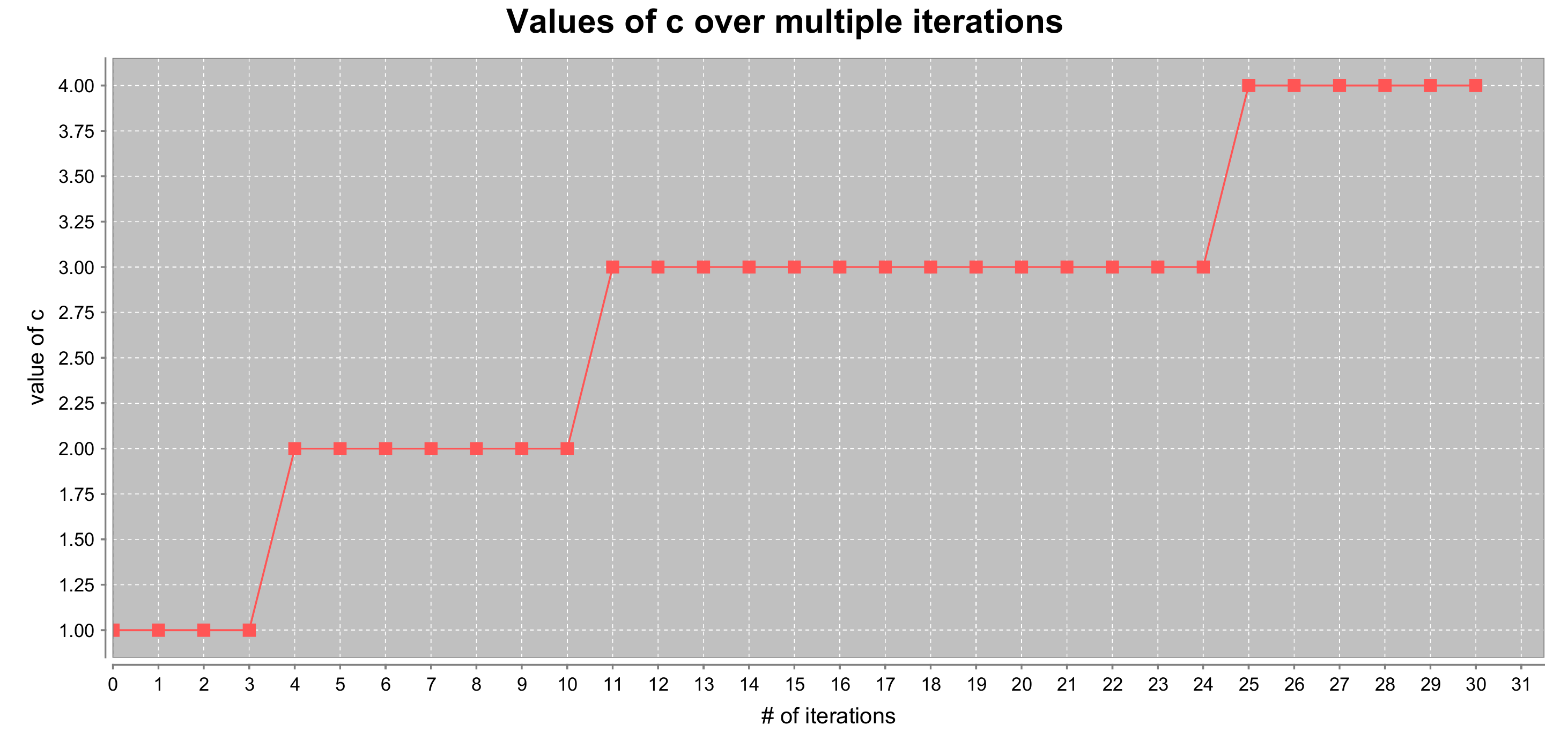}
\caption{Plot of the values of adaptor $c$ across 30 iterations of the system}
\label{PLOT_c}
\end{figure}

\section{Related Work}

A number of papers have addressed the problem of vagueness in
ontologies with a common methodology of representing the indeterminacy
of vague concepts as a static property, usually in terms of fuzzy
logic and probabilistic techniques
\cite{VagueOntologies,Stoilos:2005:FDL,Bobillo:2008:FDL,Koller:1997:PC}.  The
approach we have presented here instead situates the problem of
vagueness within the framework of ontology evolution and ontology
alignment. That is, we focus on the dynamic properties of vague
concepts, whereby there indeterminacy is bound up with their capacity
to change and adapt to different contexts.

Unlike work in Ontology Learning (e.g.,
\cite{Zhou:2007:OntologyLearning}), we are not attempting to construct
ontologies from raw sources of information, such as unstructured
text. Instead, our approach aims at computing ontological updates by
aligning an existing ontology to another source of ontological
information.

Several solutions have been proposed (e.g.,
\cite{Haase:2005:DifferentApproaches}) to the problem of resolving
inconsistencies in evolving ontologies, but none of them seem
entirely satisfactory. One option is to develop reasoning methods that
can cope with inconsistent axiom sets \cite{Ma:2007:DEO}. However,
these methods are hard to automate and can be more time-consuming than
traditional approaches. An alternative, proposed by
\cite{Heflin:2000:DOW}, is to restrict updates to those that will
preserve consistency. The disadvantage is that many kinds of ontology
update will be disallowed, including the modification of vague
concepts. Yet another strategy is to restore the consistency of an
ontology (ontology repair) when an inconsitency arises. One possibility
for automating the process of ontology repair is to remove some of the
axioms that cause the inconsistency \cite{Haase2005}. The axioms
removed, however, might roll back the new changes that were introduced
in the ontology or delete part of the ontology that should have been
preserved.  Our strategy for handling inconsistencies shares
similarities with the ontology repair approach but differs from
existing strategies in that no axioms are deleted as a result
of the repair process.

\section{Conclusion}

The \vago\ system presented here implements a novel approach for
dealing with vagueness in formal ontologies: a vague concept receives
a total interpretation (regarded as a supervaluation) but is
inherently open to change through learning. More precisely, the meaning of a vague
concept is dependent on a number of values, marked by adaptors, which
can be automatically updated. These adaptors can be used to define
cardinality restrictions and datatype range restrictions for OWL
properties.

The definitions of the vague concepts of an ontology are
automatically updated by validating the original ontology against a set of training axioms,
thereby generating an updated ontology. Inconsistencies that arise
from combining the training axioms with the the original ontology are interpreted as a
misalignment between those two sources of ontological information. This
misalignment can be reduced by modifying
the values of the adaptors used in the original ontology. 
If the axioms of another ontology are used as
training axioms for the original ontology, then the update will result
in an improved alignment between the two ontologies. The preliminary
results obtained by the simulations suggest that this framework could
be effectively used to update the definitions of vague concepts in
order to evolve a single ontology or to improve the extension-based
alignment between multiple ontologies.

\bibliographystyle{splncs03}

\bibliography{owlish}
\end{document}